\documentclass[%
twocolumn,
groupedaddress,
amsmath,amssymb, nofootinbib
]{revtex4-2}

\usepackage[english]{babel}
\usepackage{algorithm2e}
\usepackage{natbib}
\usepackage{color}

\usepackage{amsmath}
\usepackage{amssymb}
\usepackage{graphicx}
\usepackage[mathscr]{euscript}
\usepackage[colorlinks=true, allcolors=blue]{hyperref}

\begin{document}

\title{Leveraging {chaotic transients} in the training of artificial neural networks}
\author{Pedro Jiménez-González, Miguel C. Soriano and Lucas Lacasa}
\affiliation{Institute for Cross-Disciplinary Physics and Complex Systems (IFISC, CSIC-UIB),\\Campus UIB, 07122 Palma de Mallorca, Spain}

%\begin{document}

\begin{abstract}
Traditional algorithms to optimize artificial neural networks when confronted with a supervised learning task are usually exploitation-type %Miguel: this is the regime of operation for the standard parameters used in practice
relaxational dynamics such as gradient descent (GD). Here, we explore the dynamics of the neural network trajectory along training for unconventionally large learning rates. We show that for a region of values of the learning rate, the GD optimization shifts away from purely exploitation-like algorithm into a regime of exploration-exploitation balance, as the neural network is still capable of learning but the trajectory shows sensitive dependence on initial conditions --as characterized by positive network maximum Lyapunov exponent--. %The distribution of network updates in such range shares similarities with Lévy flight distributions, a paradigm of exploration-exploitation in animal search strategies.  
Interestingly, the characteristic training time required to reach an acceptable accuracy in the test set reaches a minimum precisely in such learning rate region, further suggesting that one can accelerate the training of artificial neural networks by locating at the onset of chaos. Our results --initially illustrated for the MNIST classification task--  qualitatively hold for a range of supervised learning tasks, {learning architectures (including both shallow and deep multilayer perceptrons and convolutional neural networks) and other hyperparameters (different activation functions and weight regularisation),} and showcase the emergent, constructive role of transient chaotic dynamics in the training of artificial neural networks.

\end{abstract}

\maketitle

%\vspace{6mm}
The multilayer perceptron (MLP) is an archetypical method in supervised machine learning \cite{goodfellow2016deep, aggarwal2018neural}, used to infer (regress or classify) complex input-output representations $x\to y$, where $x\in \mathbb{R}^m$ and $y$ is usually another vector for regression tasks or an element from a discrete set in classification tasks. Accordingly, an MLP is mathematically an overparametrized nonlinear function $\mathscr{F}(x;{\Omega})$, where $\Omega=\{w_k\}$ is the set of trainable parameters. 
Often represented as an (artificial neural) network \cite{goodfellow2016deep, aggarwal2018neural}, the MLP is visualized as a mathematical graph composed of various stacked layers of interconnected nodes, where the first layer represents the input vector $x\in \mathbb{R}^m$. The edges connecting nodes in adjacent layers represent (parametric) affine transformations of the vector elements of one layer to the next one (the coefficients of such transformations, usually called the edge weights, belong to the parameter set $\Omega$), and the nodes integrate the incident linear compositions in the edges nonlinearly via what is called an activation function. The information thus flows from the input layer $x$ onwards until reaching the final, output layer whose nodes represent the elements of the output $y=\mathscr{F}(x;{\Omega})$.\\
Optimizing the set of parameters $\Omega$ so that the mismatch between $\mathscr{F}(x;{\Omega})$ and $y$ --the so-called loss function $\mathscr{L}(x;\Omega)$-- is minimized is called the {\it training} process. 
In practice, training the MLP is usually done iteratively, being 
gradient descent (GD) in parameter space the gold standard, where 
\begin{equation} 
    \omega_k(t+1) = \omega_k(t) - \eta\partial_{\omega_k} \mathcal{L}(x;\Omega(t)), \ \forall \omega_k \in \Omega 
    \label{eq:GD}
\end{equation}
where $\eta$ is the so-called learning rate and $\partial_x:=\partial / \partial x$.

\medskip \noindent
Observe that, along training, the MLP is effectively performing a trajectory in the (graph) space spanned by the set of parameters $\Omega$. We contend that reinterpreting the training process as a (high-dimensional, latent) graph dynamics \cite{lacasa2022correlations} allows to inquire the inner workings of learning algorithms following physics-inspired and complexity-based epistemics \cite{nunes2024artificial, bianconi2023complex, arola2024effective}. Instead of tracking the scalar projection of the full dynamics given by the time evolution of the loss function $\mathscr{L}(x;\Omega(t))$, let us consider the actual network trajectory  \cite{danovski2024dynamical}. We argue that this change of focus --which amounts to tracking full graph trajectories of the form $(\Omega(0),\Omega(1),\Omega(2)\dots)$, where the set $\Omega(t)=\{\omega_1(t),\dots,\omega_m(t)\}$ incorporates the updated values of the so-called weights and biases of the neural architecture at training epoch $t$-- can provide valuable insights if at the same time we leverage principles and tools from network science \cite{latora2017complex, holme2012temporal, masuda2016guide, williams2022shape, badie2025initialisation, la2024deep, zheng2024learnable} and dynamical systems \cite{schuster2006deterministic, kantz2003nonlinear}.

\noindent 
Along training, it is often assumed that in a typical neural network trajectory there is an arrow of time induced by the relaxational nature of Eq.~\ref{eq:GD} which is inherited in graph space: {we often (perhaps misleadingly) assume that the dynamics always reaches equilibrium --some minimum of $\mathcal{L}(x;\Omega)$--, an intuition that
 parallels the behavior of Langevin dynamics --the continuous and stochastic counterpart to gradient descent-- which, under broad conditions, relaxes to an equilibrium distribution (typically a Boltzmann measure) over the energy landscape.}
Accordingly, GD \cite{bertsekas2003convex} is often seen as an {\it exploitation} search algorithm, that iteratively performs small improvements of an initial solution, so that some fitness function (here the loss function $\cal L$) is gradually (e.g. monotonically) decreased. This intuition is, at the end of the day, intimately related to the convergence properties of the GD scheme.
Now, discrete-time GD convergence is not always fulfilled, specially for large enough learning rates, when GD can display more exotic behavior \cite{kong2020stochasticity, herrmann2022chaotic} {(see Appendix A Figs.~\ref{fig:bifurcation} and~\ref{fig:toy_lyap} where we present a toy model of a gradient descent scheme displaying a period-doubling bifurcation cascade onto chaotic behavior, for a non-convex loss function and a sufficiently large learning rate)}. At the same time, convergence to local minima is an asymptotic behavior, and interesting dynamics often emerge in transient times \cite{lai2011transient}. Altogether, it is interesting to consider whether, led by qualitative dynamical changes in the behavior of GD-type maps, not just the loss function but the whole MLP network trajectories {\it transition} from following a pure exploitation strategy to other search strategies --such as {\it exploration} \cite{vcrepinvsek2013exploration}-- when the learning rate is large enough. 
Incidentally, note that a similar balance of strategies \cite{berger2014exploration} is known to yield optimal searching behaviors in animal foraging \cite{humphries2012foraging, ramos2004levy, eliassen2007exploration, reynolds2018levy, kembro2019bumblebees, monk2018ecology, paiva2022visibility}, transport \cite{klages2008anomalous} and a range of decision-making contexts \cite{hills2015exploration, addicott2017primer}, and is an explicit cornerstone of the Reinforcement Learning paradigm \cite{ishii2002control}.

\noindent 
Our contention in this work is that, indeed, such transition takes place, and the emergence of exploration-like dynamics,  induced by the onset of sensitive dependence on initial conditions --the hallmark of chaotic dynamics-- in the dynamics of Eq.~\ref{eq:GD} {yields a sweet spot where the balance of exploitation and exploration dynamics has a constructive, beneficial role for neural network learning}. Let us clarify at this point that in this work we are neither considering intrinsically dynamic neural networks whose neurons show chaotic behavior \cite{bertschinger2004edge, kadmon2015transition, sussillo2009generating, pereira2023forgetting, pazo2024discontinuous, luo2025butterfly} {nor considering the stability of the input-to-output trajectory in deep networks \cite{storm2024finite}}, instead we are considering the training (optimization) dynamics in a neural network whose output is not dynamic.
Moreover, we argue that (i) for an often unexplored region of large-but-not-too-large values of $\eta$, the map is optimally interpolating exploitation and exploration search strategies, and (ii) that the average training time required {for the network to learn a representation capable of generalizing --i.e. capable to reach a good performance on the test set--} $\texttt{Test} = \{x_i,y_i\}_{i=1}^{N_{\text{test}}}$ is minimized in such {\it sweet spot}. Interestingly, this coincides with a similar type of optimality emerging when the loss function's Hessian asymptotically evolves over training towards its so-called edge-of-stability, i.e. when its largest eigenvalue approaches $2/\eta$ \cite{cohen2021gradient}. 
%a condition where the maximum eigenvalue of loss function's Hessian tends to evolve within training towards
In a nutshell, we argue that the transition to an exploitation-exploration balance is achieved by leveraging the onset of transient chaotic mixing \cite{lai2011transient}, and at such transition the system efficiently minimizes training time, resulting in a possible demonstration of Langton's hypothesis \cite{langton1990computation}.

\begin{figure}[htb!]
\centering
\includegraphics[width=0.8\columnwidth]{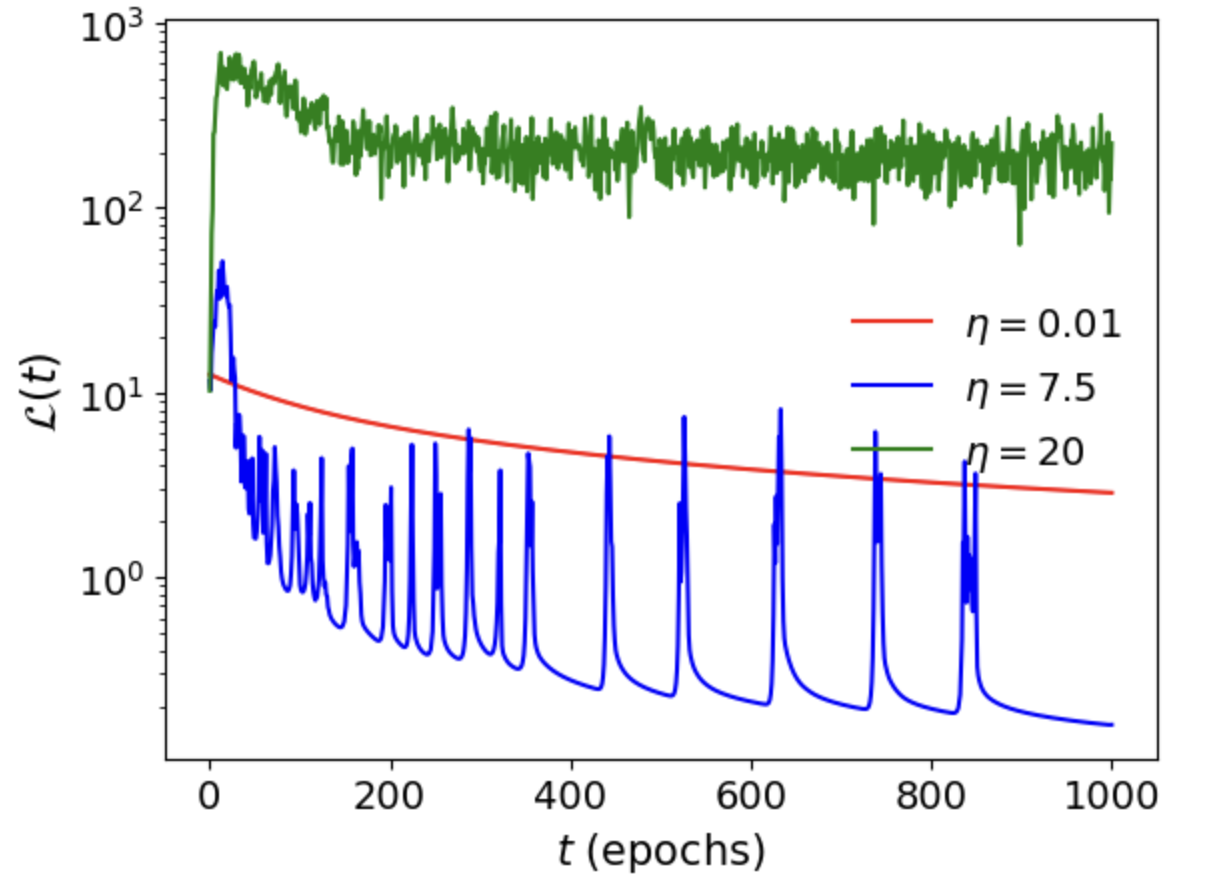}
\caption{{\small Training loss trajectory of a neural network on the MNIST dataset for three different learning rates: $\eta=0.01$, $\eta=7.5$ and $\eta=20$.}}
\label{fig:loss}
\end{figure}

\begin{figure}[htb!]
    \centering
    \includegraphics[width=0.5\textwidth]{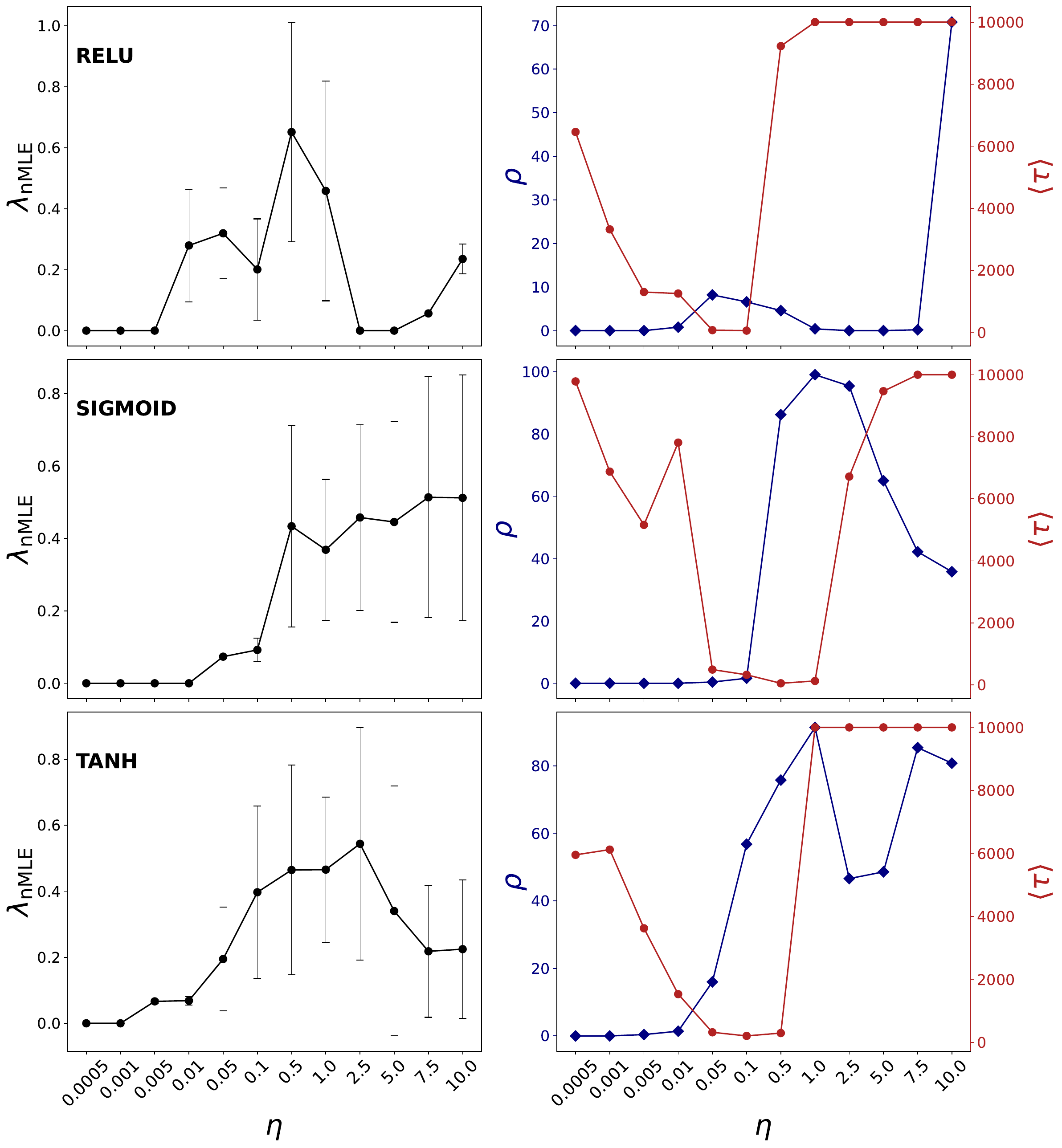}
    \caption{Lyapunov exponent (left column), percentage of valid exponents ($\rho$) and mean convergence time ($\langle \tau \rangle$) as functions of the learning rate ($\eta$) for each activation function. Results for the Iris task.}
    \label{fig:iris}
\end{figure}

\begin{figure}[htb!]
    \centering
    \includegraphics[width=0.5\textwidth]{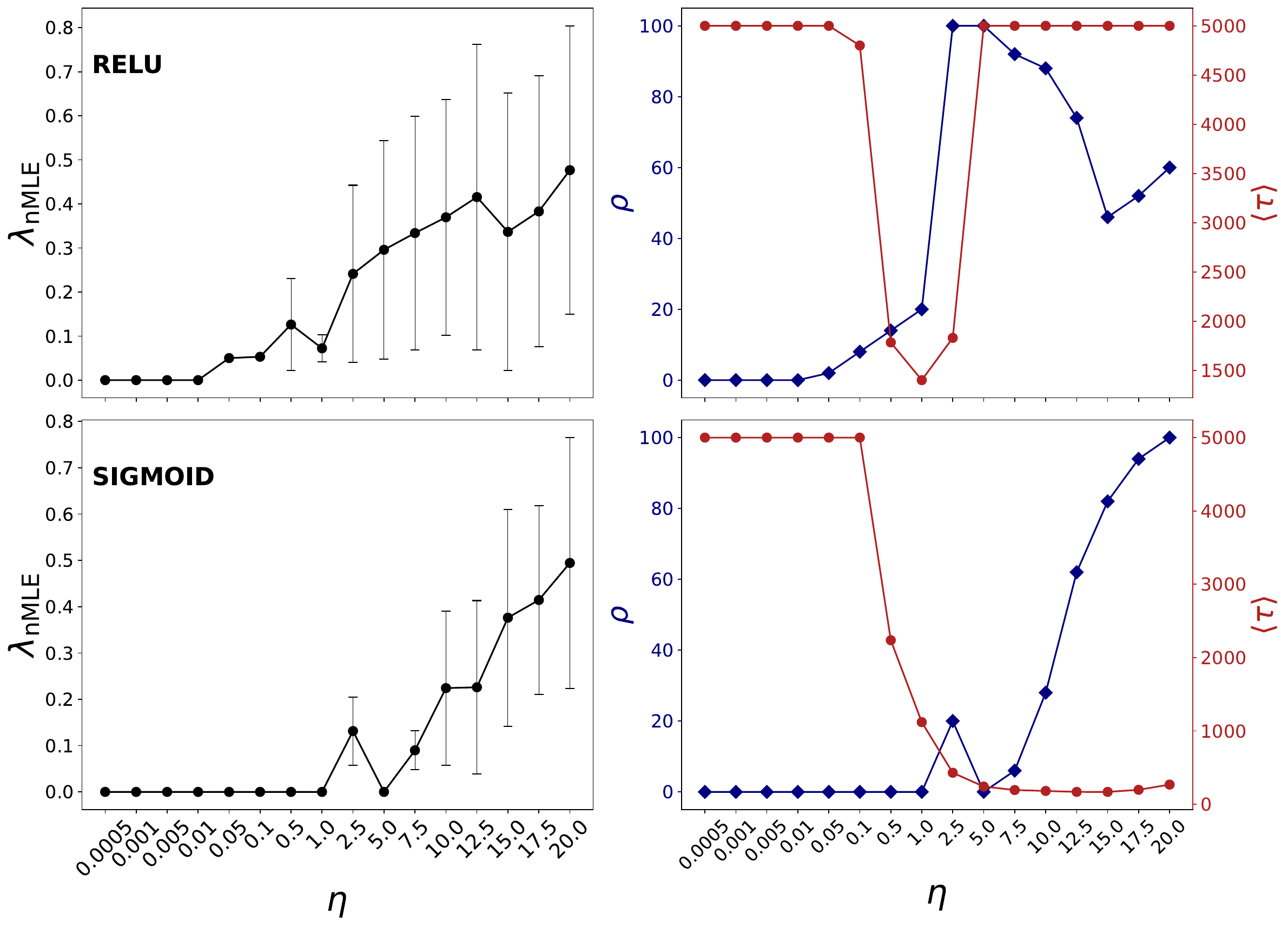}
    \caption{Lyapunov exponent (left column), percentage of valid exponents ($\rho$) and mean convergence time ($\langle \tau \rangle$) as functions of the learning rate ($\eta$) for the remaining activation functions. Results for the MNIST task.}
    \label{fig:mnist}
\end{figure}

\begin{figure}[htb!]
    \centering
    \includegraphics[width=0.5\textwidth]{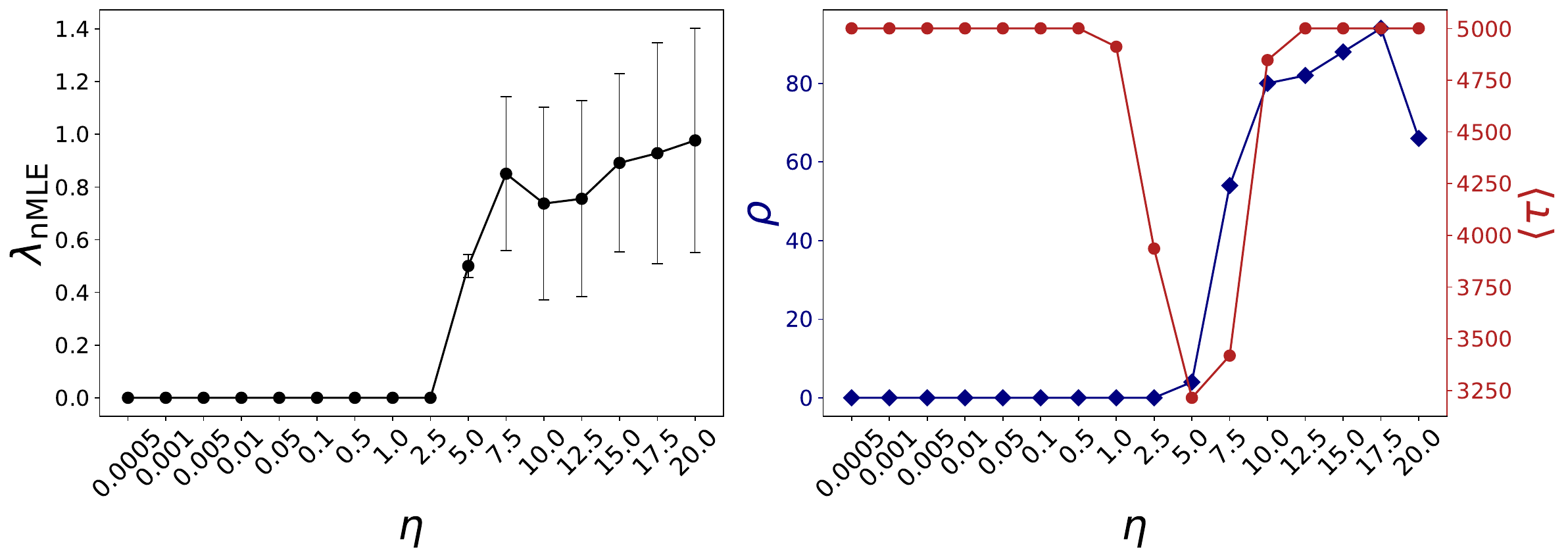} 
    \caption{Lyapunov exponent (left column), percentage of valid exponents ($\rho$) and mean convergence time ($\langle \tau \rangle$) as functions of the learning rate ($\eta$). Results for the MNIST task with an extra hidden layer in the architecture and tanh as the activation function.}
    \label{fig:MNIST_extra}
\end{figure}

\begin{figure}[htb!]
    \centering
    \includegraphics[width=0.5\textwidth]{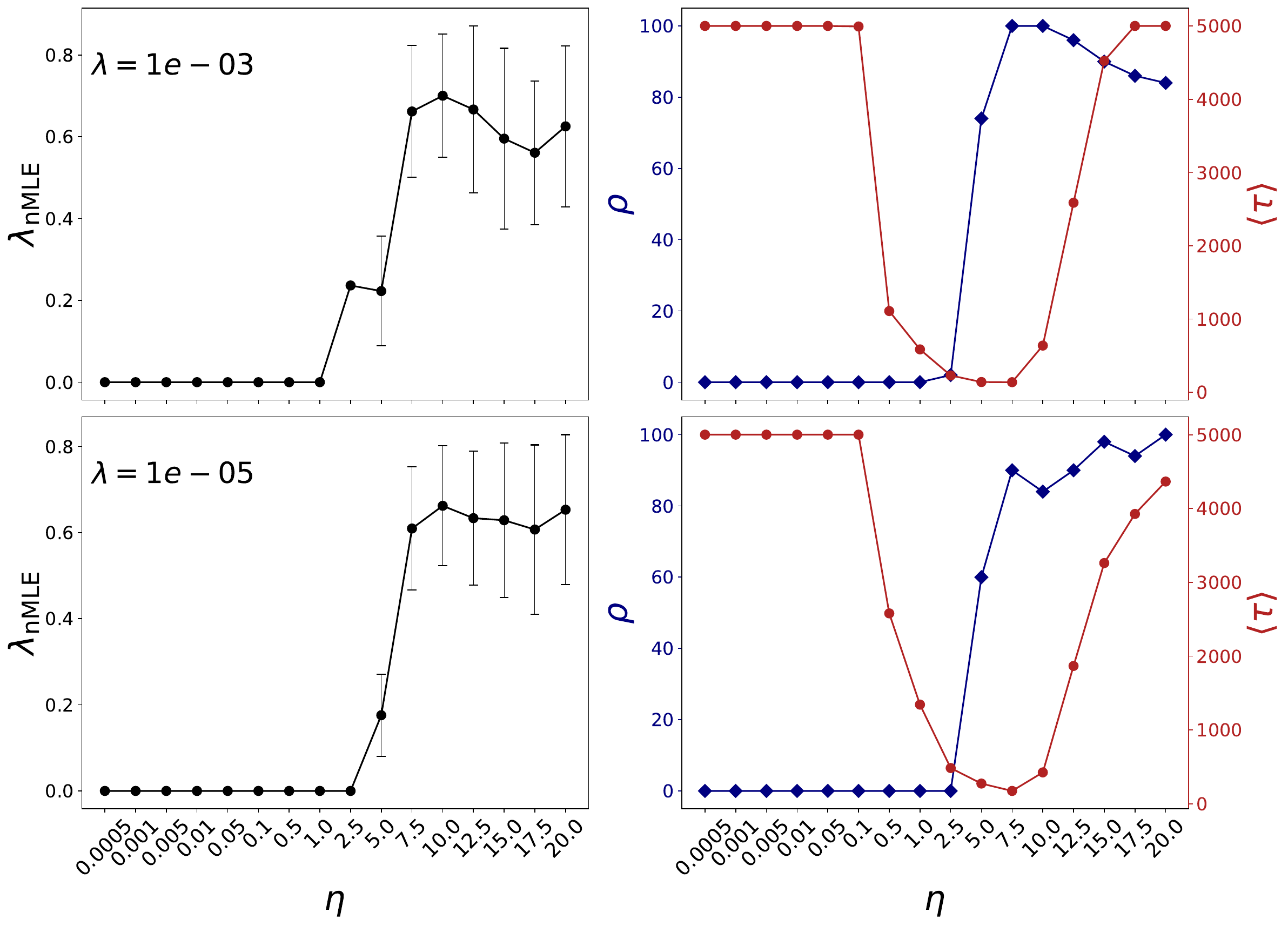}
    \caption{Lyapunov exponent (left column), percentage of valid exponents ($\rho$) and mean convergence time ($\langle \tau \rangle$) as functions of the learning rate ($\eta$). Results for the MNIST task with a single hidden layer and tanh activation function, now including $L_{2}$-regularization.}
    \label{fig:MNIST_regularization}
\end{figure}

\begin{figure}[htb!]
    \centering
    \includegraphics[width=0.5\textwidth]{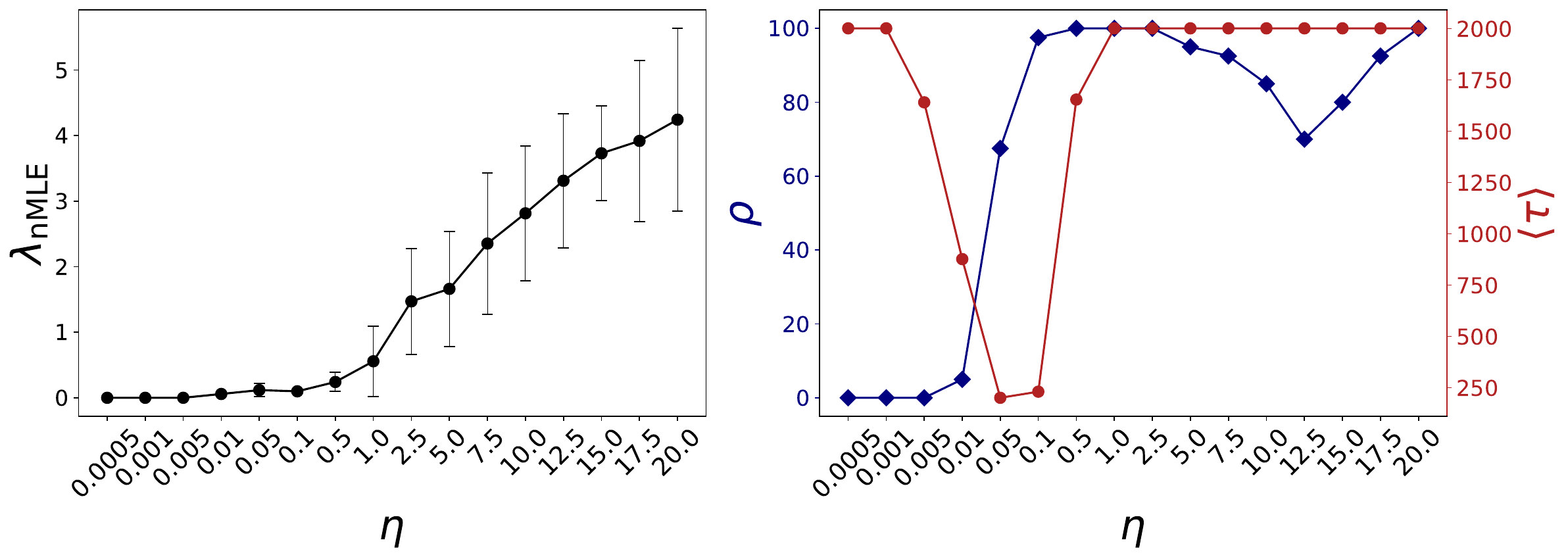}
    \caption{Lyapunov exponent (left column), percentage of valid exponents ($\rho$) and mean convergence time ($\langle \tau \rangle$) as functions of the learning rate ($\eta$). Results correspond to the CNN applied to the MNIST classification task.}
    \label{fig:result_cnn}
\end{figure}

\begin{figure}[htb!]
    \centering
    \includegraphics[width=0.5\textwidth]{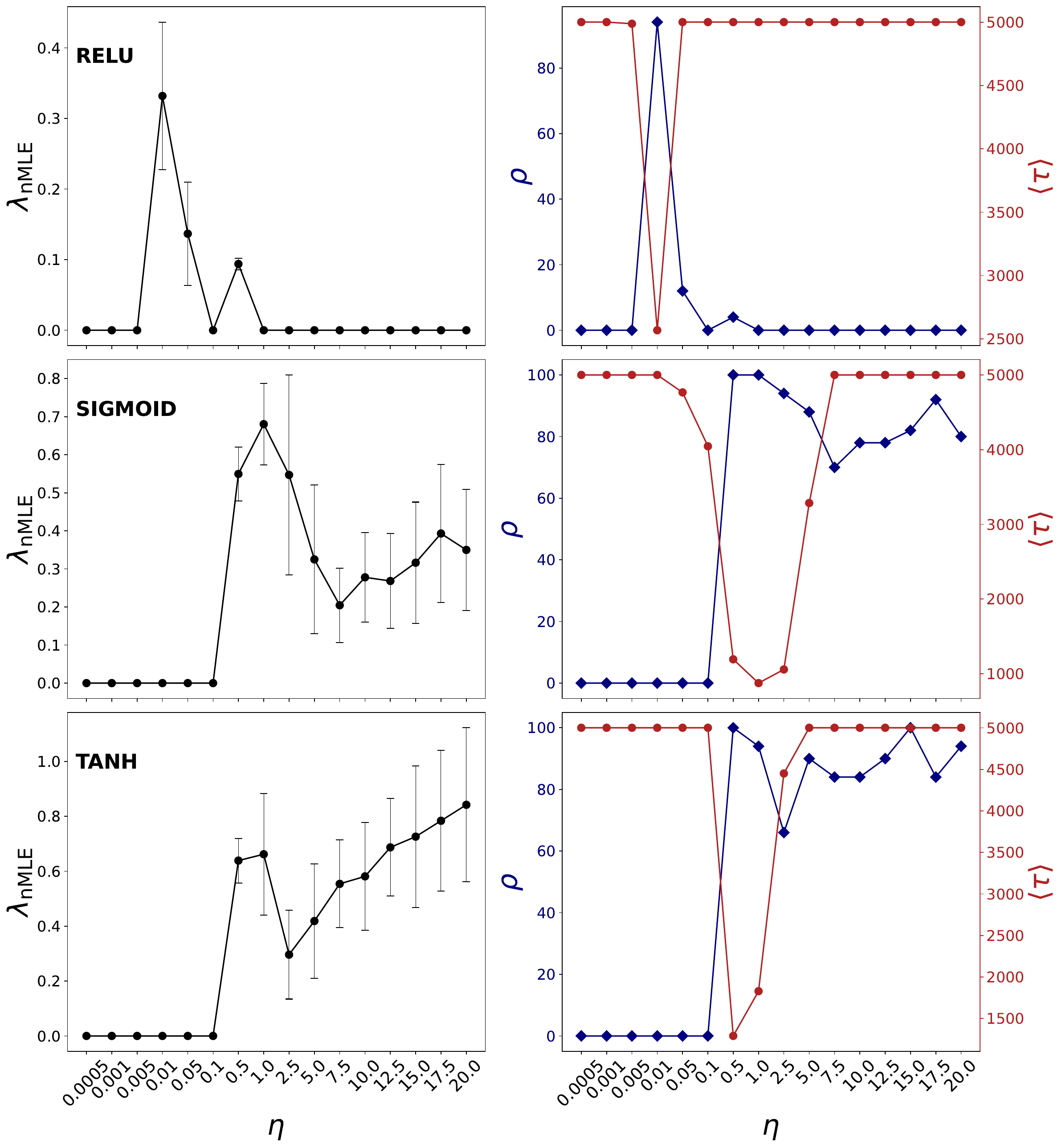}
    \caption{Lyapunov exponent (left column), percentage of valid exponents ($\rho$) and mean convergence time ($\langle \tau \rangle$) as functions of the learning rate ($\eta$) for each activation function. Results for the CIFAR-10 task with a target accuracy of 30\%.}
    \label{fig:cifar}
\end{figure}

\noindent 
To actually demonstrate our hypothesis, we initially consider a vanilla classification task: MNIST image classification \cite{lecun1998mnist}. For parsimony, we initially choose a shallow MLP  (one hidden layer with 64 neurons) and a $\tanh$ activation function (the results for other supervised learning tasks, activation functions deeper architectures and other architectures with inductive biases are discussed in the {Appendix B Figs.~\ref{fig:iris}--\ref{fig:cifar}}). After train/test split, this MLP is trained on a  set $\texttt{Train}=\{(x_i,y_i)\}_{i=1}^N$ of $N={6\cdot 10^4}$ labelled handwritten images ($y_i$ is the label of the $i$-th image), and we use a cross-entropy loss function
\begin{equation}
    \mathcal{L}(x;\Omega)=-\frac{1}{N}\sum_{i=1}^Ny_{i}\log \mathscr{F}(x_{i};\Omega).\label{loss}
\end{equation}
Training takes place by using Eq.~\ref{eq:GD}. For the sake of simplicity, no regularization is initially added to the loss function (results with $L_2$ weight regularization are discussed in {Appendix B Fig.~\ref{fig:MNIST_regularization}}). Additionally, we use traditional GD schemes (no mini-batch or SGD) and discard dropout, to remove any source of stochasticity to the network dynamics \cite{kong2020stochasticity}. Finally, to assess the performance for different learning rates, a constant learning rate $\eta$ is fixed throughout training.

\medskip
\noindent Fig.~\ref{fig:loss} plots the time evolution of the (training) loss function $\mathcal{L}(x;\Omega(t))\equiv {\cal L}(t)$ for three different learning rates $\eta$. Interestingly, such loss is only monotonically decreasing for the standard range of small values of the learning rate. For larger values of $\eta$ other dynamical behaviors are found: converging loss functions with non-monotonic, irregular transients and other dynamical attractors for extremely large $\eta$, where the MLP does not seem to make any useful learning.

\begin{figure}[htb]
\centering
\includegraphics[width=0.8\linewidth]{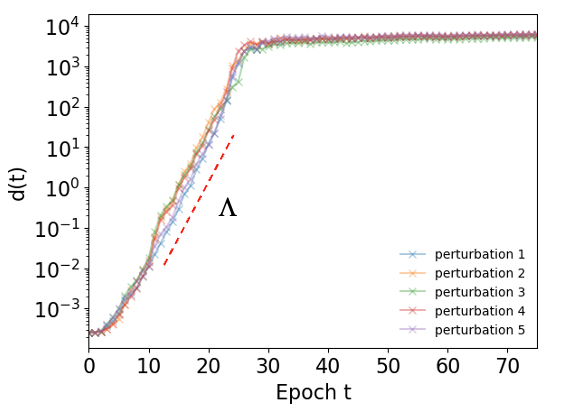}
\caption{{\small Semi-log plot of the evolution (along training) of the network distance $d(t)$ for pairs of network trajectories with closeby initialization $\Omega(0)$, as a function of the number of epochs $t$, for a shallow MLP with $\tanh()$ activation function trained on MNIST with a large learning rate. $d(t)$ displays a stylized exponential expansion followed by saturation. The slope of the exponential phase corresponds to the local network Lyapunov exponent $\Lambda$ and is indicative of chaotic mixing.}}
\label{fig:d}
\end{figure}

\noindent
To better characterize what particular change in dynamical behaviors in the network evolution is inducing these projections in the loss function, we now resort to recently-introduced graph-theoretical extensions of the Maximum Lyapunov Exponent \cite{annalisa, caligiuri2025characterizing}, designed to estimate sensitivity to initial conditions in network trajectories. The procedure consists in three steps: (i) for a fixed learning rate, we define a set of $q$ different network initializations ${\cal S}= \{\Omega(0)\}$. (ii) Around each concrete network initialization $\Omega(0)$, we build an $\epsilon$-ball formed by a set of $M$ small network `perturbations' of ${\cal B}=\{\Omega(0)^{(j)}\}_{j=1}^M$. To do that, we perturb each trainable parameter $w_k \in \Omega$ with uniform noise $w_k' = w_k + \xi$, $\xi \sim \textsc{Uniform}(-\epsilon, \epsilon)$. Defining the distance between two network initializations $\Omega, \Omega'$ as the $L_1$ norm $d(\Omega, \Omega')=\sum_{\omega_k \in \Omega} |\omega_k - \omega_k'|$, it follows that each perturbed network $\Omega(0)^{(j)}$ is at most at distance $\textsc{Card}(\Omega)\cdot \epsilon$ from $\Omega(0)$.
(iii) Then, following \cite{annalisa} we measure the expansion rate of closeby network trajectories by adequately averaging the divergence of the $M$ elements inside each $\epsilon$-ball throughout the action of the training dynamics in Eq.~\ref{eq:GD}:
\begin{equation}\label{eq:lyapunov}
    \Lambda_{\Omega(0)}=\frac{1}{\tau}\ln\frac{M^{-1}\sum_{j=1}^{M}d_{j}(\tau)}{M^{-1}\sum_{j=1}^{M}d_{j}(0)},
\end{equation}
where $d_j(\tau)\equiv d(\Omega(\tau),\Omega(\tau)^{(j)})$.
$\Lambda_{\Omega(0)}$ is indeed the network version of a finite, local Lyapunov exponent \cite{annalisa}, where $\tau$ is the characteristic time required for the elements inside the $\epsilon$-ball centered at $\Omega(0)$ to diverge to distances of the order of the phase space diameter. For an illustration, in Fig.~\ref{fig:d} we depict the distance $d(t)$ between $M=5$ network initializations close to an initial condition $\Omega(0)$ ($\epsilon = 10^{-8}$), for a large learning rate $\eta=10$. We observe a clear exponential phase up to $\tau\approx 30$ epochs, the slope denoting the finite local network Lyapunov exponent $\Lambda_{\Omega(0)}$.

\begin{figure}[htb!]
\centering
\includegraphics[width=0.8\linewidth]{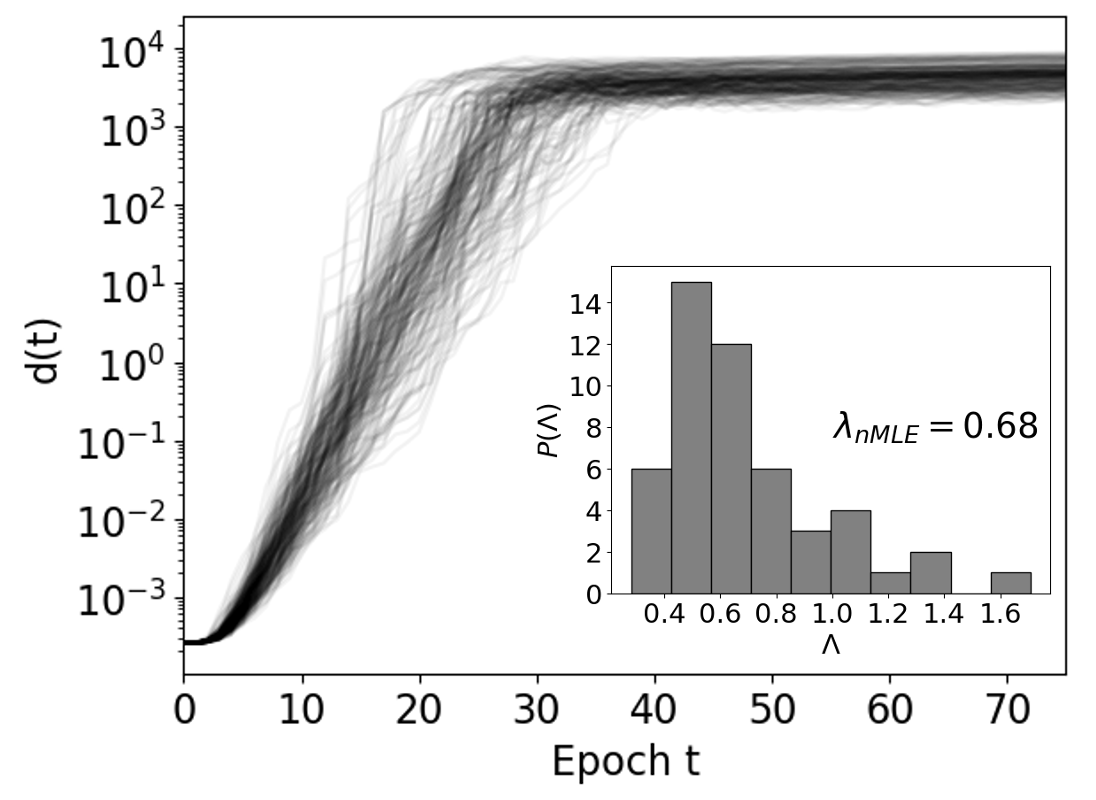}
\caption{{\small Same as Fig~\ref{fig:d}, but for many different $\epsilon$-balls centered at different initial conditions $\Omega$. Each initial condition leads in principle to a different local network Lyapunov exponent $\Lambda(\Omega)$. In the inset, we display the histogram of local network Lyapunov exponents. The average of this distribution is the estimation of the network MLE $\lambda_{\text{nMLE}}\approx 0.68$.}}
\label{fig:MLE}
\end{figure}

\medskip
\noindent 
The network's Maximum Lyapunov exponent averages the local exponents over different network initializations $\lambda_{\text{nMLE}}=\langle \Lambda_{\Omega(0)} \rangle_{\Omega(0)\in {\cal S}}$. For illustration, this is displayed for $\textsc{Card}({\cal S})=50$ different initial conditions in Fig.~\ref{fig:MLE}. The inset of that panel depicts the histogram of local exponents, whose average gives $\lambda_{\text{nMLE}}\approx 0.68$, i.e. the system shows sensitivity to initial conditions.
%\footnote{Note that for the exponential expansion to take place already from $t=0$, the initial condition needs to be in the chaotic attractor, something which occurs with vanishing probability for dissipative dynamics. Instead, distance $d(t)$ evolves more or less constantly for $t<t_{\text{trans}}$, i.e. for a few epochs until the trajectories settle into the chaotic attractor, and then deviate exponentially fast according to $\lambda_{\text{nMLE}}$. In the example displayed in Fig.~\ref{fig:MLE}, $t_{\text{trans}}$ is very small, but for other architectures and tasks, $t_{\text{trans}}$ can be larger.}.\\ 
$\lambda_{\text{nMLE}}$ can thus be seen as an order parameter distinguishing two phases: a phase where the search strategy induced by the GD map Eq.~\ref{eq:GD} is an exploitation one and $\lambda_{\text{nMLE}} \le 0$, and a phase where the search strategy is of an exploration type with sensitive dependence on initial conditions, and $\lambda_{\text{nMLE}} > 0$. The transition between both phases interpolates both search strategies.
As a complementary metric that will later assists us in our empirical analysis of the network training dynamics, we also
%Inspired by the fluctuation-dissipation relation in statistical physics, we also compute the variance $\sigma^2(\Lambda)=\langle \Lambda^2\rangle_{\Omega} - \langle \Lambda \rangle_{\Omega}^2$ --analog to the system's susceptibility--. 
define $\rho$ as the percentage of MLP initializations of the set $\cal S$ for which the $\epsilon$-ball expansion can be well approximated by an exponential function with high statistical significance ($R^2>0.9$, $\Lambda > 0.05$), i.e
this is the percentage of local network Lyapunov exponents which are positive.
%i.e. when when the 
%exponential fit
%$p$-value of fitting $\sum_{j=1}^{M}d_{j}(k)$ to an exponential function is smaller than $0.01$.\\

\begin{figure}[htb!]
\centering
\includegraphics[width=0.8\linewidth]{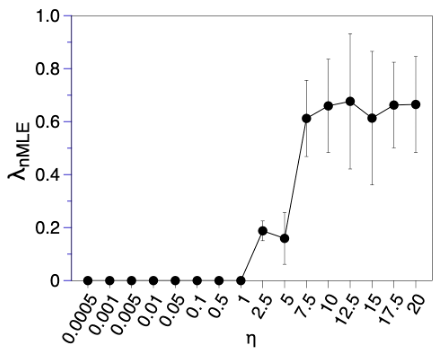}
\caption{{\small Estimation of the network Maximum Lyapunov Exponent $\lambda_{\text{nMLE}}$ for MLP trajectories as a function of the learning rate $\eta$. Error bars denote $\pm$ one standard deviation of the population of finite local network Lyapunov exponents $\{\Lambda(\Omega)\}$. The onset of sensitivity to initial conditions $\lambda_{\text{nMLE}}>0$ marks the change from a purely exploitation-type optimization to an exploration/exploitation type.}}
\label{fig:MLE_vs_eta}
\end{figure}

\begin{figure}[htb!]
\centering
\includegraphics[width=0.8\linewidth]{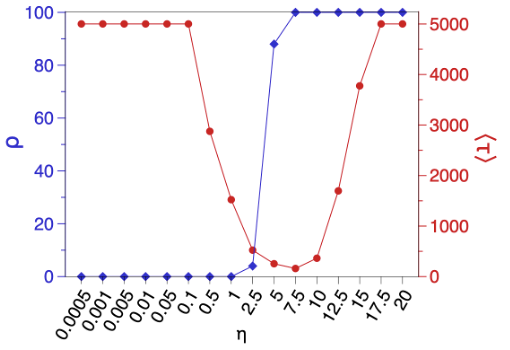}
\caption{{\small Blue diamonds depict $\rho$, the percentage of MLP initializations $\Omega$ leading to training trajectories with positive local Lyapunov exponent $\Lambda(\Omega)>0$ as a function of the learning rate $\eta$. In the same figure, we also plot (red dots) the average training time $\langle \tau \rangle$ (in number of Gradient Descent epochs) needed to reach an accuracy of 90\% or larger in the test set. Training is found to be maximally efficient close to the onset of fully-developed sensitivity to initial conditions ($\Lambda(\Omega)>0 \ \forall \Omega$).}}
\label{fig:rho_tau}
\end{figure}

\noindent
In Fig.~\ref{fig:MLE_vs_eta} we report $\lambda_{\text{nMLE}}$ as a function of the gradient descent's learning rate $\eta$. Results indicate that there is a clear transition between exploitation to exploration search, where the exploitation-exploration interpolation hovers in the range $\eta \in [1,10]$. Fig~\ref{fig:rho_tau} displays $\rho$ as a function of the learning rate. This metric confirms a transition between a phase where no initial conditions display exponential expansion, to a phase where virtually all regions of the phase space display chaotic transients.\\
\noindent Finally, to assess the MLP's learning and training efficiency in the context of the abovementioned phenomenology, Fig.~\ref{fig:rho_tau} also depicts the average number of epochs $\langle \tau \rangle$ needed for the MLP to reach an average classification accuracy of at least $0.9$ in the test set (this is averaged over all different MLP initializations in $\cal S$), as a function of the learning rate $\eta$. We find that $\langle  \tau \rangle$ displays a non-monotonic shape, and indeed reaches a minimum in the exploitation-exploration interpolation region, at the learning rate $\eta \approx 7.5$, precisely marking the onset of fully-developed sensitivity to initial conditions $\rho \approx 100\%$.\\
As advanced, this type of optimality in the training dynamics has been recently observed when the largest eigenvalue $\sigma_{\text{max}}$ of the loss function's Hessian asymptotic converges to $2/\eta$, the so-called edge of stability \cite{cohen2021gradient}. In Appendix C Fig.~\ref{fig:sharpness}, we show how the time series of $\sigma_{\text{max}}(t)$ over training indeed approaches $2/\eta$ when the learning rate hovers around the values for which $\langle  \tau \rangle$ is minimized, and suggests that asymptotically self-organizing around the Hessian's edge-of-stability is precursed by a chaotic transient.

\medskip \noindent Our results are robust against changes in the classification task (see Appendix B Fig.~\ref{fig:iris} and~\ref{fig:cifar} for further results on the Iris and CIFAR-10 classification respectively), the type of activation function (see Appendix B Fig.~\ref{fig:mnist} for further results with sigmoid and ReLU functions), 
the depth of the MLP (see Appendix B Fig.~\ref{fig:MNIST_extra} for a comparison of shallow vs deep network), the inclusion of weight regularization (see Appendix B Fig.~\ref{fig:MNIST_regularization}), and {also hold for other neural network architectures with inductive biases (see Appendix B Fig.~\ref{fig:result_cnn} for analysis on a convolutional neural network)} and overall highlight the relevance of leveraging {transient} chaotic mixing in the training of neural networks. In summary, we have found that, as the learning rate increases, the training dynamics transition from a regular, purely exploitation-type dynamics  to a chaotic, purely exploration-type dynamics. The transition between both types is rather sharp and occurs in a region that trades-off exploitation and exploration by the onset of a mechanism of {chaotic transient --a chaotic fingerprint emerging in the first few dozen epochs of the training dynamics--}. This mechanism underpins an efficient search of graph space, eventually leading to faster learning. Evidence suggests that such early-stage, chaotic transient is precursing at a later Hessian trajectories to eventually converge towards their edge of stability.

\medskip \noindent From a conceptual point of view, our findings suggest a demonstration of Langton's edge of chaos hypothesis \cite{langton1990computation}, and a confirmation of Verschure's seminal idea of using chaos as a fast-search mechanism \cite{verschure1991chaos}. {In hindsight, this is an example where instabilities emerging at the core of numerical schemes --often seen as a nuisance when it comes to e.g. integrating differential equations or root finding-- have in the case of searching for neural network representations an unexpected benefit. This benefit occurs because in neural network optimization one uses GD as a {\it search} algorithm rather than simply as a local minimizer, and as a consequence, exploitation benefits from exploration when both mechanisms are at play. Aside from this theoretical insight, forcing the training dynamics' learning rate to be located at the sweet spot could also yield practical benefits. This can be achieved e.g. by} 
%we argue that this phenomenology could be leveraged to automatically boost the training efficiency of MLPs. As a matter of fact, while a priori the optimeal learning rate might depend on the specific task and architecture, results suggest that the existence of such optimal learning rate is universally valid. Accordingly, and since $\langle \tau \rangle$ is  substantially reduced at such learning rate, one could e.g. 
{using the bisection method to find the sweet spot, by iteratively refining an initial learning rate range $[\eta_{\text{min}}, \eta_{\text{max}}]$, such that $\rho(\eta_{\text{min}})\approx 0$ and $\rho(\eta_{\text{max}})\approx 100$ as a pre-processing step, before actually training the system.}\\
{Finally, further work is needed to clarify the modulating effect of stochastic sources in the network trajectories, such as using stochastic or mini-batch as opposed to full-batch gradient descent \cite{herrmann2022chaotic, dandi2024two} or adding dropout, and to address more sophisticated learning rate schedules. Likewise, investigating whether such instability-induced advantage holds in other optimization schemes, or whether other parameters such as the batch size \cite{dandi2024two} could play the role of a control parameter, and exploring if such phenomenology can be retrieved from low-dimensional projections of network trajectories \cite{lacasa2025scalar, lacasa2025eigendecompositions}, are all interesting open problems.}

\medskip
\noindent {\bf Acknowledgments --} The authors thank K. Danovski for input in preliminary stages of this project, M. Matías for insightful discussions, A. Fernández Amil for pointing us to \cite{verschure1991chaos} after our work was submitted as a preprint and referees for stimulating discussions.
PJ ackowledges funding from Maria de Maeztu (MdM) Seal of Excellence  (CEX2021-001164-M) via the FPI programme (grant PRE2022-104148), funded by the MICIU/AEI/10.13039/501100011033.
LL acknowledges partial support from project CSxAI (PID2024-157526NB-I00) funded by MICIU/AEI/10.13039/501100011033/FEDER, UE. MCS and LL acknowledge partial support from a Maria de Maeztu grant (CEX2021-001164-M) funded by the MICIU/AEI/10.13039/501100011033, and from the European Commission Chips Joint Undertaking project No. 101194363 (NEHIL).\\

\medskip
%\appendix
%\section{Appendix A: Emergence of chaotic dynamics in toy gradient descent maps.} 
\setcounter{equation}{0}
\renewcommand{\theequation}{A\arabic{equation}}
\noindent {\it Appendix A: Emergence of chaotic dynamics in toy gradient descent maps.} Often, gradient descent optimisation of neural networks weights generates monotonically decreasing loss function trajectories towards one of the minima, and common wisdom assumes that such monotonicity is also inherited in weight space. Here we challenge this belief by presenting a toy model for the (gradient descent) optimisation of neural networks which display chaotic trajectories in weight space.\\
Let us consider a one-dimensional gradient dynamics map
\begin{equation}
    w_{t+1}=w_t - \eta \frac{d V(w)}{dw}\equiv f(w_t),
    \label{eq:toy}
\end{equation}
where the learning rate $\eta$ is a map's parameter and $V(w)$ is a potential function that acts as our simplified loss function. The simplest case of a non-convex loss with multiple minima is perhaps a double well potential
\begin{equation}
    V(w) = \frac{1}{4}w^4 - \frac{1}{2}w^2,
    \label{eq:potential}
\end{equation}
which has two local minima for $w=-1,1$. In this case, $f(w)=w[1-\eta(w^2-1)]$ is thus a cubic map. We now show that even in such simple setting, the dynamics can become chaotic for sufficiently large learning rates.

\begin{figure}[h]
    \centering
    \includegraphics[width=0.5\textwidth]{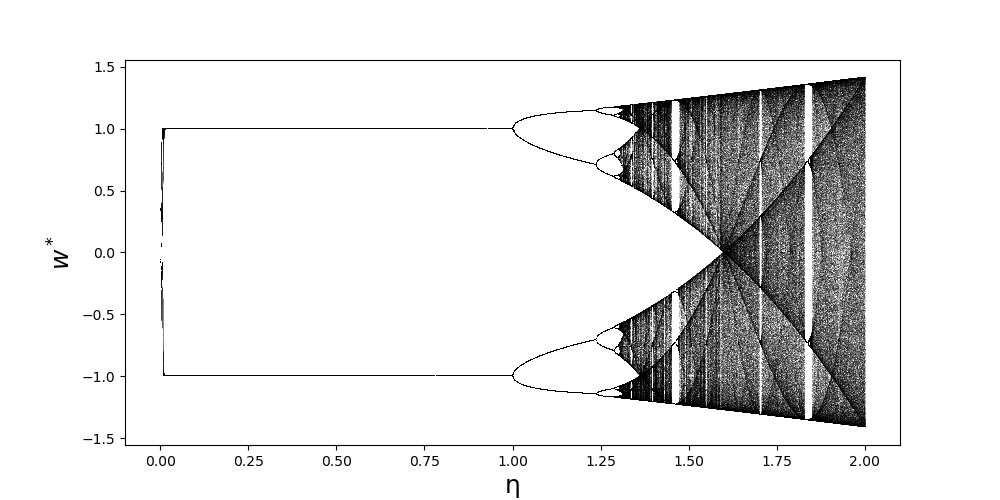}
    \caption{{Bifurcation diagram of the one-dimensional gradient map, displaying a period-doubling cascade to chaos.}}
    \label{fig:bifurcation}
\end{figure}

\medskip
While for low values of $\eta$ the dynamics of Eq.~\ref{eq:toy} converge towards one of the two minima, for larger values of this parameter, a bifurcation cascade takes place. In Fig.~\ref{fig:bifurcation} we plot the bifurcation diagram of the map, obtained numerically by iterating Eq.~\ref{eq:toy} for 200 steps a random initial conditions $w_0\in N(0,0.1)$ (after discarding a transient period) as a function of $\eta$. The figure shows a classical period-doubling bifurcation cascade towards a chaotic state with aperiodic dynamics, intertwined with some self-affine periodic windows. This is the first evidence.\\
Since this map is one-dimensional, it only has a single Lyapunov exponent $\lambda$, which can be approximated up to arbitrary precision with
\begin{equation}
    \lambda = \lim_{t\to \infty }\frac{1}{t}\sum_{k=0}^{t-1} |\ln(f'(w_k)|.
\end{equation}
In Fig.~\ref{fig:toy_lyap} we show $\lambda$ as a function of the learning rate $\eta$, clearly finding regions for which $\lambda>0$. All in all, this toy model illustrates that gradient descent scheme can yield chaotic dynamics (in the sense of Devaney) for non-convex loss functions (such as the ones involved in neural network optimisation) for sufficiently large learning rates. Since realistic (deep) neural network optimisation just involves substantially more non-convex loss functions of substantially higher-dimensional weight vectors, it is reasonable to expect that moving towards this more complex scenario makes the onset of chaotic behavior for large learning rates a generic phenomenon.

\begin{figure}[h]
    \centering
    \includegraphics[width=0.5\textwidth]{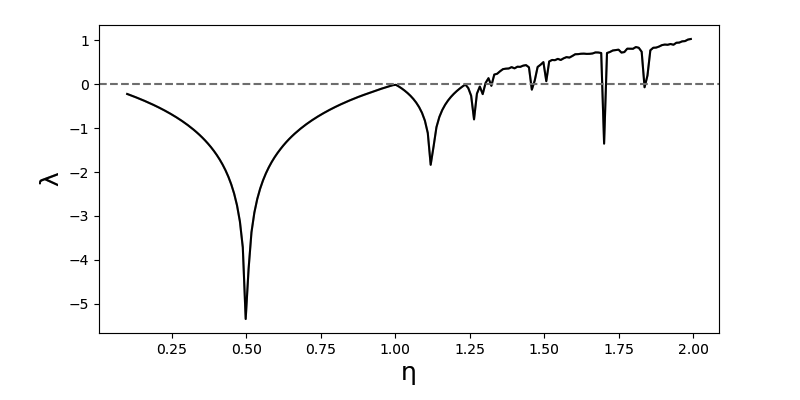}
    \caption{{Lyapunov exponent of the gradient map in Eq.~\ref{eq:toy} as a function of the learning rate $\eta$, showing regions with chaotic behavior.}}
    \label{fig:toy_lyap}
\end{figure}

% Pedro: en principio quedamos en suprimir aquí la parte de Task Datasets, different architectures, activation functions and L2 regularization. Unificar esto en un appendix Experimental details más simple y conciso?

\medskip
\noindent {\it Appendix B: Onset of sensitivity to initial conditions and training efficiency: Supplementary results.} Below we present the results of the key metrics (network Maximum Lyapunov Exponent $\lambda_\text{nMLE}$, percentage of positive finite network Lyapunov exponents $\rho$, and average training time $\langle \tau \rangle$) for the additional set of configurations and tasks (Iris vs MNIST vs CIFAR-10, different activation functions, shallow vs deep architecture, effect of regularization and CNNs). These results provide additional evidence for the robustness of the phenomenology described in the main text. {Unless otherwise stated, all the networks are initialized with weights drawn from $\mathcal{N}(0,1)$ and biases initialized to zero, and are trained using the same protocol as in the main text. A baseline implementation of the data-generation pipeline used in this work is publicly available at \cite{code}.}

\medskip \noindent

%We start with the Iris task and present the results in Fig.~\ref{fig:iris}, following the same methodology as in the main part of the paper. The results obtained using the Iris dataset illustrate the behavior of the training network dynamics in a low-dimensional case. Due to its small size and low computational cost, Iris provides an accessible testbed to validate the methodology before scaling it to more complex datasets. Results are qualitatively similar to the ones found in the main part of the paper.

{We start with the Iris classification task \cite{irisdataset} (150 samples, 4 features, 3 classes; 120/30 train/test split). The network architecture is 4–10–3 (83 trainable parameters). Results are shown in Fig.~\ref{fig:iris}. Despite the low dimensionality of the task, the same qualitative transition is observed: the minimum in $\langle \tau \rangle$ coincides with the onset of positive $\lambda_{\mathrm{nMLE}}$.}

\medskip \noindent
%In Fig.~\ref{fig:mnist} we come back to the MNIST task and present the results for different activation functions, highlighting the robustness of the patterns observed across different datasets and hyperparameters. For the sigmoid activation function, it should be noted that the minimum in $\langle \tau \rangle$ extends over a wide range of learning rates, forming a relatively flat region. To again observe the rise in the training time, you have to reach a learning rate value of $\eta=30$. 
{We next consider the MNIST dataset \cite{lecun1998mnist} (60,000/10000 train/test samples; 28$\times$28 grayscale images) using the 784-64-10 architecture (50,890 parameters). Figure~\ref{fig:mnist} reports results for sigmoid and ReLU activation functions, complementing the tanh case analyzed in the main text. The qualitative structure persists across nonlinearities. For the sigmoid case, the minimum of $\langle \tau \rangle$ extends over a broad range of learning rates, with the subsequent increase appearing only for $\eta \gtrsim 30$.}

\medskip \noindent %The experiments with a deep architecture are presented in Fig.~\ref{fig:MNIST_extra}, showing qualitatively similar phenomenology than the one found for the shallow architecture. In this case, the minimum number of epochs needed to reach the target accuracy is increased by a factor of 20 compared to the shallow configuration. Nevertheless, the phenomenon persists: there is still a clear minimum corresponding to the region of $\eta$ where $\lambda_{nMLE}$ starts to take positive values.
{To assess depth effects, we replace the single hidden layer (64 neurons) by two hidden layers of 32 neurons each (784–32–32–10; 26,506 parameters), keeping total hidden width fixed. Results in Fig.~\ref{fig:MNIST_extra} show that the phenomenology persists, although the minimum convergence time increases by approximately a factor of 20 relative to the shallow configuration.}

\medskip \noindent
%Now we present the results of the simulations with the inclusion of $L_{2}$ weight regularization in Fig.~\ref{fig:MNIST_regularization}. The regularization strength $\lambda$ was set to $10^{-3}$ and $10^{-5}$, chosen as representative values commonly used in practice to explore the influence of soft and weak regularisation effects. We also see that a suitable region appears if we look at the range of learning rate values that simultaneously minimise the training time of the network and exhibit chaotic dynamics.
{We further evaluate $L_{2}$-regularized training with loss $\mathcal{L}_\lambda = \mathcal{L} + \frac{\lambda}{2}\|\mathscr{W}\|^2$ using $\lambda = 10^{-3}$ and $10^{-5}$ for the MNIST–tanh configuration. As shown in Fig.~\ref{fig:MNIST_regularization}, the region of optimal training efficiency remains aligned with the onset of positive $\lambda_{\mathrm{nMLE}}$.}

\medskip \noindent
%To conclude the analysis on the MNIST dataset, Fig.~\ref{fig:result_cnn} presents the results obtained with the CNN architecture. The observed phenomenon persists, exhibiting a sweet spot for the learning rate which remains evident even with a neural network architecture that includes a clear inductive bias for image processing. To adapt to architectural changes and computational limitations, the simulation parameters were modified: the CNN was trained for a maximum of 2,000 epochs, with 40 different initial conditions and a target accuracy of 80\%.
{Finally, we consider a convolutional architecture for MNIST consisting of two convolutional layers (8 and 16 kernels, $3\times3$, padding=1), each follow by $2\times2$ max pooling, and a fully connected output layer (9,098 trainable parameters). Convolutional layers use ReLU activations and weights are initialized from $\mathcal{N}(0,0.01)$. Training is performed for up to 2,000 epochs with 40 initial conditions and a target accuracy of 80\%. As shown in Fig.~\ref{fig:result_cnn}, the characteristic sweet spot in learning rate remains visible.}

\medskip \noindent %Finally we present the results obtained for the CIFAR-10 task in Fig.~\ref{fig:cifar}. Although the prediction task is not successfully solved, since it is not the main focus of our study, we include this case to demonstrate that the observed phenomenon persists. Networks with this type of topology (MLP) typically do not achieve accuracies higher than 50\% on CIFAR-10. Achieving accuracies of approximately 70\% usually requires more sophisticated MLP architectures, such as including the use of linear bottleneck layers \cite{lin2015far}. For this reason, we lower the accuracy threshold to 30\% and observe that the same characteristic behaviour is maintained.
{We conclude with the CIFAR-10 dataset \cite{cifar10dataset} (50,000/10,000 train/test samples; 32$\times$32 RGB images) using the 3072–256–10 MLP architecture (789,258 parameters). Since MLPs typically achieve limited performance on this task, we set the target accuracy to 30\%. As shown in Fig.~\ref{fig:cifar}, the same qualitative behavior is observed despite the increased task complexity.}

\medskip 
\noindent {\it Appendix C: Evolution of sharpness.}
Following the approach in \cite{cohen2021gradient}, we measure the evolution of sharpness during training, defined as the maximum eigenvalue of the Hessian of the training loss with respect to the model parameters. This measure characterises the local curvature of the loss surface and provides insight into the dynamics of the optimisation process. The sharpness is tracked across epochs for the model trained with the MNIST dataset and tanh as activation function.\\
We use power iteration to approximate the top eigenvalue of the Hessian. This method is computationally efficient and does not require explicit computation and storage of the full Hessian matrix, so it is feasible to apply it during training.
We track the evolution of the sharpness over training by measuring the top eigenvalue of the Hessian matrix of the loss function, as shown in Figure \ref{fig:sharpness}. For learning rates that minimize $\langle \tau \rangle$, we find that, in line with theoretical considerations \cite{cohen2021gradient}, sharpness remains bounded once the training has stabilised and closely matches the theoretical limit of $2/\eta$, as indicated by the red dashed line. The results shown correspond to learning rates $\eta=7.5$ and $\eta=10$.

\begin{figure}[htb!]
    \centering
    \includegraphics[width=0.5\textwidth]{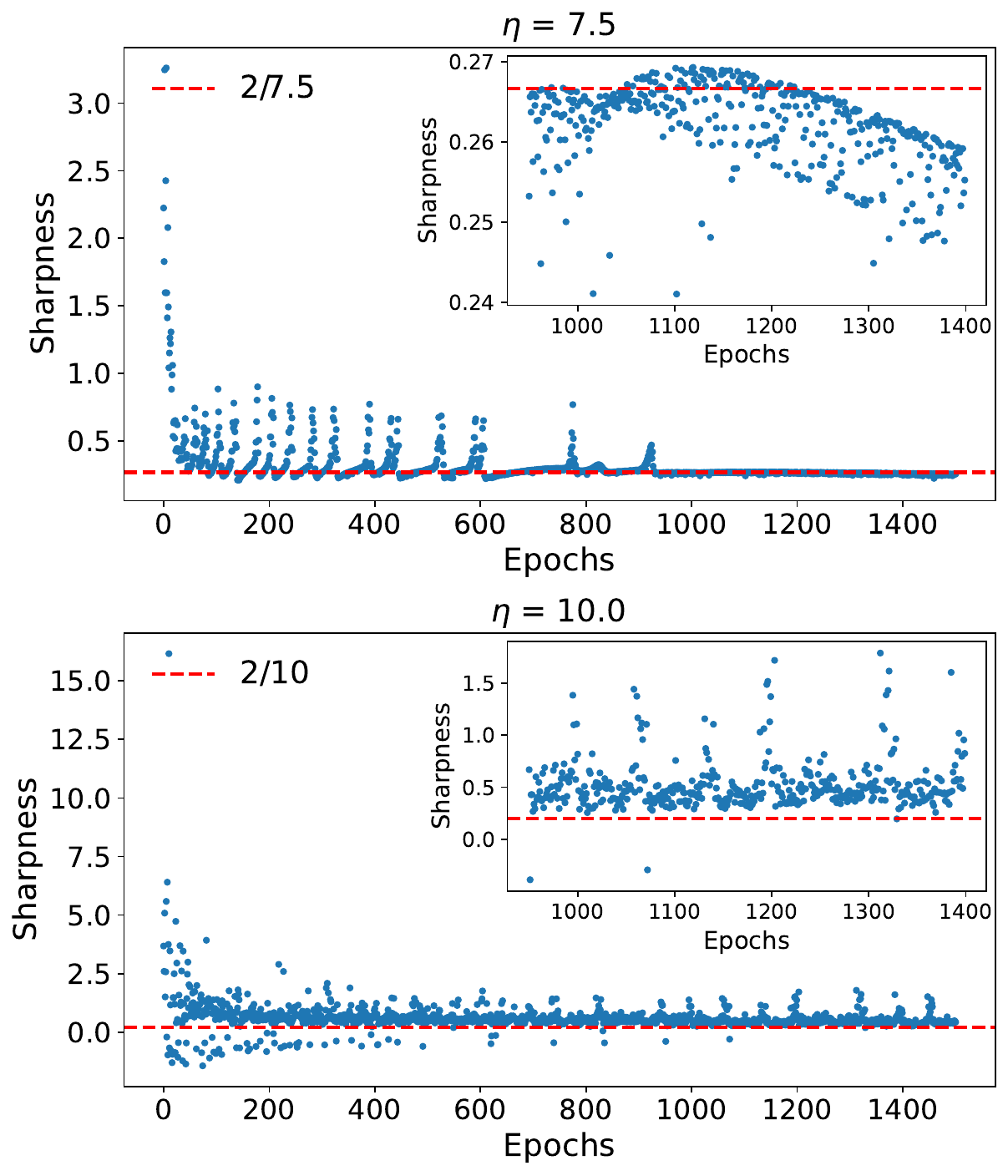}
    \caption{Evolution of sharpness as a function of training epochs. The theoretical value $2/\eta$ is indicated by the dashed red line for reference. Results are shown for learning rates values $\eta=7.5$ and $\eta=10$. }
    \label{fig:sharpness}
\end{figure}

%\bibliographystyle{unsrt}
%\bibliography{sample}

\begin{thebibliography}{10}

\bibitem{goodfellow2016deep}
Ian~J. Goodfellow, Yoshua Bengio, and Aaron Courville.
\newblock {\em Deep Learning}.
\newblock MIT Press, Cambridge, MA, USA, 2016.
\newblock \url{http://www.deeplearningbook.org}.

\bibitem{aggarwal2018neural}
Charu~C Aggarwal.
\newblock {\em Neural networks and deep learning}, volume~10.
\newblock Springer, 2018.

\bibitem{lacasa2022correlations}
Lucas Lacasa, Jorge~P Rodriguez, and Victor~M Eguiluz.
\newblock Correlations of network trajectories.
\newblock {\em Physical Review Research}, 4(4):L042008, 2022.

\bibitem{nunes2024artificial}
Lu{\'\i}s~A Nunes~Amaral.
\newblock Artificial intelligence needs a scientific method-driven reset.
\newblock {\em Nature Physics}, 20(4):523--524, 2024.

\bibitem{bianconi2023complex}
Ginestra Bianconi, Alex Arenas, Jacob Biamonte, Lincoln~D Carr, Byungnam Kahng,
  Janos Kertesz, J{\"u}rgen Kurths, Linyuan L{\"u}, Cristina Masoller,
  Adilson~E Motter, et~al.
\newblock Complex systems in the spotlight: next steps after the 2021 nobel
  prize in physics.
\newblock {\em Journal of Physics: Complexity}, 4(1):010201, 2023.

\bibitem{arola2024effective}
Llu{\'\i}s Arola-Fern{\'a}ndez and Lucas Lacasa.
\newblock Effective theory of collective deep learning.
\newblock {\em Physical Review Research}, 6(4):L042040, 2024.

\bibitem{danovski2024dynamical}
Kaloyan Danovski, Miguel~C Soriano, and Lucas Lacasa.
\newblock Dynamical stability and chaos in artificial neural network
  trajectories along training.
\newblock {\em Frontiers in Complex Systems}, 2:1367957, 2024.

\bibitem{latora2017complex}
Vito Latora, Vincenzo Nicosia, and Giovanni Russo.
\newblock {\em Complex networks: principles, methods and applications}.
\newblock Cambridge University Press, 2017.

\bibitem{holme2012temporal}
Petter Holme and Jari Saram{\"a}ki.
\newblock Temporal networks.
\newblock {\em Physics reports}, 519(3):97--125, 2012.

\bibitem{masuda2016guide}
Naoki Masuda and Renaud Lambiotte.
\newblock {\em A guide to temporal networks}.
\newblock World Scientific, 2016.

\bibitem{williams2022shape}
Oliver~E Williams, Lucas Lacasa, Ana~P Mill{\'a}n, and Vito Latora.
\newblock The shape of memory in temporal networks.
\newblock {\em Nature communications}, 13(1):499, 2022.

\bibitem{badie2025initialisation}
Arash Badie-Modiri, Chiara Boldrini, Lorenzo Valerio, J{\'a}nos Kert{\'e}sz,
  and M{\'a}rton Karsai.
\newblock Initialisation and network effects in decentralised federated
  learning.
\newblock {\em Applied Network Science}, 10(1):53, 2025.

\bibitem{la2024deep}
Emanuele La~Malfa, Gabriele La~Malfa, Giuseppe Nicosia, and Vito Latora.
\newblock Deep neural networks via complex network theory: a perspective.
\newblock {\em arXiv preprint arXiv:2404.11172}, 2024.

\bibitem{zheng2024learnable}
Ziwei Zheng, Huizhi Liang, Vaclav Snasel, Vito Latora, Panos Pardalos, Giuseppe
  Nicosia, and Varun Ojha.
\newblock On learnable parameters of optimal and suboptimal deep learning
  models.
\newblock {\em arXiv preprint arXiv:2408.11720}, 2024.

\bibitem{schuster2006deterministic}
Heinz~Georg Schuster and Wolfram Just.
\newblock {\em Deterministic chaos: an introduction}.
\newblock John Wiley \& Sons, 2006.

\bibitem{kantz2003nonlinear}
Holger Kantz and Thomas Schreiber.
\newblock {\em Nonlinear time series analysis}.
\newblock Cambridge university press, 2003.

\bibitem{bertsekas2003convex}
Dimitri Bertsekas, Angelia Nedic, and Asuman Ozdaglar.
\newblock {\em Convex analysis and optimization}, volume~1.
\newblock Athena Scientific, 2003.

\bibitem{kong2020stochasticity}
Lingkai Kong and Molei Tao.
\newblock Stochasticity of deterministic gradient descent: Large learning rate
  for multiscale objective function.
\newblock {\em Advances in neural information processing systems},
  33:2625--2638, 2020.

\bibitem{herrmann2022chaotic}
Luis Herrmann, Maximilian Granz, and Tim Landgraf.
\newblock Chaotic dynamics are intrinsic to neural network training with sgd.
\newblock {\em Advances in Neural Information Processing Systems},
  35:5219--5229, 2022.

\bibitem{lai2011transient}
Ying-Cheng Lai and Tam{\'a}s T{\'e}l.
\newblock {\em Transient chaos: complex dynamics on finite time scales}, volume
  173.
\newblock Springer Science \& Business Media, 2011.

\bibitem{vcrepinvsek2013exploration}
Matej {\v{C}}repin{\v{s}}ek, Shih-Hsi Liu, and Marjan Mernik.
\newblock Exploration and exploitation in evolutionary algorithms: A survey.
\newblock {\em ACM computing surveys (CSUR)}, 45(3):1--33, 2013.

\bibitem{berger2014exploration}
Oded Berger-Tal, Jonathan Nathan, Ehud Meron, and David Saltz.
\newblock The exploration-exploitation dilemma: a multidisciplinary framework.
\newblock {\em PloS one}, 9(4):e95693, 2014.

\bibitem{humphries2012foraging}
Nicolas~E Humphries, Henri Weimerskirch, Nuno Queiroz, Emily~J Southall, and
  David~W Sims.
\newblock Foraging success of biological l{\'e}vy flights recorded in situ.
\newblock {\em Proceedings of the National Academy of Sciences},
  109(19):7169--7174, 2012.

\bibitem{ramos2004levy}
Gabriel Ramos-Fern{\'a}ndez, Jos{\'e}~L Mateos, Octavio Miramontes, Germinal
  Cocho, Hern{\'a}n Larralde, and Barbara Ayala-Orozco.
\newblock L{\'e}vy walk patterns in the foraging movements of spider monkeys
  (ateles geoffroyi).
\newblock {\em Behavioral ecology and Sociobiology}, 55:223--230, 2004.

\bibitem{eliassen2007exploration}
Sigrunn Eliassen, Christian J{\o}rgensen, Marc Mangel, and Jarl Giske.
\newblock Exploration or exploitation: life expectancy changes the value of
  learning in foraging strategies.
\newblock {\em Oikos}, 116(3):513--523, 2007.

\bibitem{reynolds2018levy}
Andy Reynolds, Eliane Ceccon, Cristina Baldauf, Tassia Karina~Medeiros, and
  Octavio Miramontes.
\newblock L{\'e}vy foraging patterns of rural humans.
\newblock {\em PLoS one}, 13(6):e0199099, 2018.

\bibitem{kembro2019bumblebees}
Jackelyn~M Kembro, Mathieu Lihoreau, Joan Garriga, Ernesto~P Raposo, and
  Frederic Bartumeus.
\newblock Bumblebees learn foraging routes through exploitation--exploration
  cycles.
\newblock {\em Journal of the Royal Society Interface}, 16(156):20190103, 2019.

\bibitem{monk2018ecology}
Christopher~T Monk, Matthieu Barbier, Pawel Romanczuk, James~R Watson, Josep
  Al{\'o}s, Shinnosuke Nakayama, Daniel~I Rubenstein, Simon~A Levin, and Robert
  Arlinghaus.
\newblock How ecology shapes exploitation: a framework to predict the
  behavioural response of human and animal foragers along
  exploration--exploitation trade-offs.
\newblock {\em Ecology letters}, 21(6):779--793, 2018.

\bibitem{paiva2022visibility}
Leticia~R Paiva, Sidiney~G Alves, Lucas Lacasa, Og~DeSouza, and Octavio
  Miramontes.
\newblock Visibility graphs of animal foraging trajectories.
\newblock {\em Journal of Physics: Complexity}, 3(4):04LT03, 2022.

\bibitem{klages2008anomalous}
Rainer Klages, G{\"u}nter Radons, and Igor~Mihajlovi{\v{c}} Sokolov.
\newblock {\em Anomalous transport}.
\newblock Wiley Online Library, 2008.

\bibitem{hills2015exploration}
Thomas~T Hills, Peter~M Todd, David Lazer, A~David Redish, and Iain~D Couzin.
\newblock Exploration versus exploitation in space, mind, and society.
\newblock {\em Trends in cognitive sciences}, 19(1):46--54, 2015.

\bibitem{addicott2017primer}
Merideth~A Addicott, John~M Pearson, Maggie~M Sweitzer, David~L Barack, and
  Michael~L Platt.
\newblock A primer on foraging and the explore/exploit trade-off for psychiatry
  research.
\newblock {\em Neuropsychopharmacology}, 42(10):1931--1939, 2017.

\bibitem{ishii2002control}
Shin Ishii, Wako Yoshida, and Junichiro Yoshimoto.
\newblock Control of exploitation--exploration meta-parameter in reinforcement
  learning.
\newblock {\em Neural networks}, 15(4-6):665--687, 2002.

\bibitem{bertschinger2004edge}
Nils Bertschinger, Thomas Natschl{\"a}ger, and Robert Legenstein.
\newblock At the edge of chaos: Real-time computations and self-organized
  criticality in recurrent neural networks.
\newblock {\em Advances in neural information processing systems}, 17, 2004.

\bibitem{kadmon2015transition}
Jonathan Kadmon and Haim Sompolinsky.
\newblock Transition to chaos in random neuronal networks.
\newblock {\em Physical Review X}, 5(4):041030, 2015.

\bibitem{sussillo2009generating}
David Sussillo and Larry~F Abbott.
\newblock Generating coherent patterns of activity from chaotic neural
  networks.
\newblock {\em Neuron}, 63(4):544--557, 2009.

\bibitem{pereira2023forgetting}
Ulises Pereira-Obilinovic, Johnatan Aljadeff, and Nicolas Brunel.
\newblock Forgetting leads to chaos in attractor networks.
\newblock {\em Physical Review X}, 13(1):011009, 2023.

\bibitem{pazo2024discontinuous}
Diego Paz{\'o}.
\newblock Discontinuous transition to chaos in a canonical random neural
  network.
\newblock {\em Physical Review E}, 110(1):014201, 2024.

\bibitem{luo2025butterfly}
Jingyi Luo, Jianyu Chen, and Hong-Kun Zhang.
\newblock The butterfly effect in neural networks: Unveiling hyperbolic chaos
  through parameter sensitivity.
\newblock {\em Neural Networks}, page 107572, 2025.

\bibitem{storm2024finite}
L~Storm, Hampus Linander, J~Bec, Kristian Gustavsson, and Bernhard Mehlig.
\newblock Finite-time lyapunov exponents of deep neural networks.
\newblock {\em Physical Review Letters}, 132(5):057301, 2024.

\bibitem{cohen2021gradient}
Jeremy Cohen, Simran Kaur, Yuanzhi Li, J~Zico Kolter, and Ameet Talwalkar.
\newblock Gradient descent on neural networks typically occurs at the edge of
  stability.
\newblock In {\em International Conference on Learning Representations}, 2021.

\bibitem{langton1990computation}
Chris~G Langton.
\newblock Computation at the edge of chaos: Phase transitions and emergent
  computation.
\newblock {\em Physica D: nonlinear phenomena}, 42(1-3):12--37, 1990.

\bibitem{lecun1998mnist}
Yann LeCun.
\newblock The mnist database of handwritten digits.
\newblock {\em http://yann. lecun. com/exdb/mnist/}, 1998.

\bibitem{annalisa}
Annalisa Caligiuri, Victor~M. Egu\'{\i}luz, Leonardo Di~Gaetano, Tobias Galla,
  and Lucas Lacasa.
\newblock Lyapunov exponents for temporal networks.
\newblock {\em Phys. Rev. E}, 107:044305, Apr 2023.

\bibitem{caligiuri2025characterizing}
Annalisa Caligiuri, Tobias Galla, and Lucas Lacasa.
\newblock Characterizing the dynamics of unlabeled temporal networks.
\newblock {\em Chaos: An Interdisciplinary Journal of Nonlinear Science},
  35(5), 2025.

\bibitem{verschure1991chaos}
PFMJ Verschure.
\newblock Chaos-based learning.
\newblock {\em Complex Systems}, 5(S 359):370, 1991.

\bibitem{dandi2024two}
Yatin Dandi, Florent Krzakala, Bruno Loureiro, Luca Pesce, and Ludovic Stephan.
\newblock How two-layer neural networks learn, one (giant) step at a time.
\newblock {\em Journal of Machine Learning Research}, 25(349):1--65, 2024.

\bibitem{lacasa2025scalar}
Lucas Lacasa, F~Javier Mar{\'\i}n-Rodr{\'\i}guez, Naoki Masuda, and Llu{\'\i}s
  Arola-Fern{\'a}ndez.
\newblock Scalar embedding of temporal network trajectories.
\newblock {\em Chaos, Solitons \& Fractals}, 199:116599, 2025.

\bibitem{lacasa2025eigendecompositions}
Lucas Lacasa.
\newblock Fluid dynamics meet network science: two cases of temporal network
  eigendecomposition.
\newblock {\em arXiv preprint arXiv:2509.03135}, 2025.

\bibitem{code}
\url{https://github.com/pedrojg8/chaotic-transients-in-ANNs}.

\bibitem{irisdataset}
Ronald~A Fisher.
\newblock The use of multiple measurements in taxonomic problems.
\newblock {\em Annals of eugenics}, 7(2):179--188, 1936.

\bibitem{cifar10dataset}
A.~Krizhevsky and G.~Hinton.
\newblock Learning multiple layers of features from tiny images.
\newblock {\em Master's thesis, Department of Computer Science, University of
  Toronto}, 2009.

\end{thebibliography}

\end{document}